\documentclass[10pt,a4paper]{article}

\usepackage{voxprog_preprint,times}

\usepackage{graphicx}
\usepackage{booktabs}
\usepackage{bm}
\usepackage{amsmath}
\usepackage{amssymb}
\usepackage{multirow}
\usepackage{threeparttable}
\usepackage{array}
\usepackage{colortbl}
\usepackage{enumitem}
\usepackage{pifont}
\usepackage{wrapfig}
\usepackage{xspace}
\usepackage[most]{tcolorbox}
\usepackage[hidelinks]{hyperref}
\usepackage{url}
\usepackage{placeins}
\usepackage{needspace}
\usepackage[font=small,skip=4pt]{caption}
\usepackage{fancyvrb,tikz}
\usetikzlibrary{arrows.meta,positioning}

\tcbset{
 apicard/.style={
 colback=black!3,
 colframe=black!35,
 boxrule=0.4pt,
 arc=1mm,
 left=5pt,
 right=5pt,
 top=4pt,
 bottom=4pt,
},
}

\newcolumntype{I}{!{\vrule width 1pt}}

\definecolor{bblue}{rgb}{0.000,0.588,0.902}
\definecolor{mygray}{gray}{.9}
\definecolor{rui}{rgb}{0.482,0.408,0.933}
\definecolor{ggray}{rgb}{0.498,0.498,0.498}
\definecolor{movedobject}{HTML}{0086B3}
\definecolor{referenceobject}{HTML}{D35400}

\newcommand{\sssection}[1]{\par\Needspace{3\baselineskip}\noindent\textbf{#1}\enspace\ignorespaces}

\DeclareRobustCommand{\method}{\texorpdfstring{\mbox{\textsc{\fontfamily{ptm}\selectfont VoxProg}}}{VoxProg}\xspace}
\DeclareRobustCommand{\suite}{\texorpdfstring{\mbox{\textsc{\fontfamily{ptm}\selectfont VoxProg}}-Bench}{VoxProg-Bench}\xspace}

\title{
\method: 3D Spatial Reasoning through Voxel Programming}

\author{
 {\sffamily\bfseries
\mbox{Zhichao Zhai\textsuperscript{1}},
\mbox{Menglei Zhao\textsuperscript{1}},
\mbox{Guikun Chen\textsuperscript{1}},
\mbox{Zihao Yan\textsuperscript{2}},\par
\mbox{Kanzhi Wu\textsuperscript{2}},
\mbox{Wenguan Wang\textsuperscript{1}},
\mbox{Jun Xiao\textsuperscript{1,\ensuremath{\dagger}}}\par}
\vspace{5pt}
\textsuperscript{1}Zhejiang University\par
\textsuperscript{2}vivo Mobile Communication Co., Ltd., Shenzhen, China

}

\hypersetup{pdftitle={VoxProg: 3D Spatial Reasoning through Voxel Programming},pdfauthor={Zhichao Zhai, Menglei Zhao, Guikun Chen, Zihao Yan, Kanzhi Wu, Wenguan Wang, Jun Xiao}}
\begin{document}
\raggedbottom
\setlength{\parskip}{4pt}
\maketitle
\begingroup
\renewcommand{\thefootnote}{\ensuremath{\dagger}}
\footnotetext[2]{Corresponding author: Jun Xiao.}
\endgroup
\suppressfloats[t]

\begin{figure}[t]
\centering
\includegraphics[width=\linewidth]{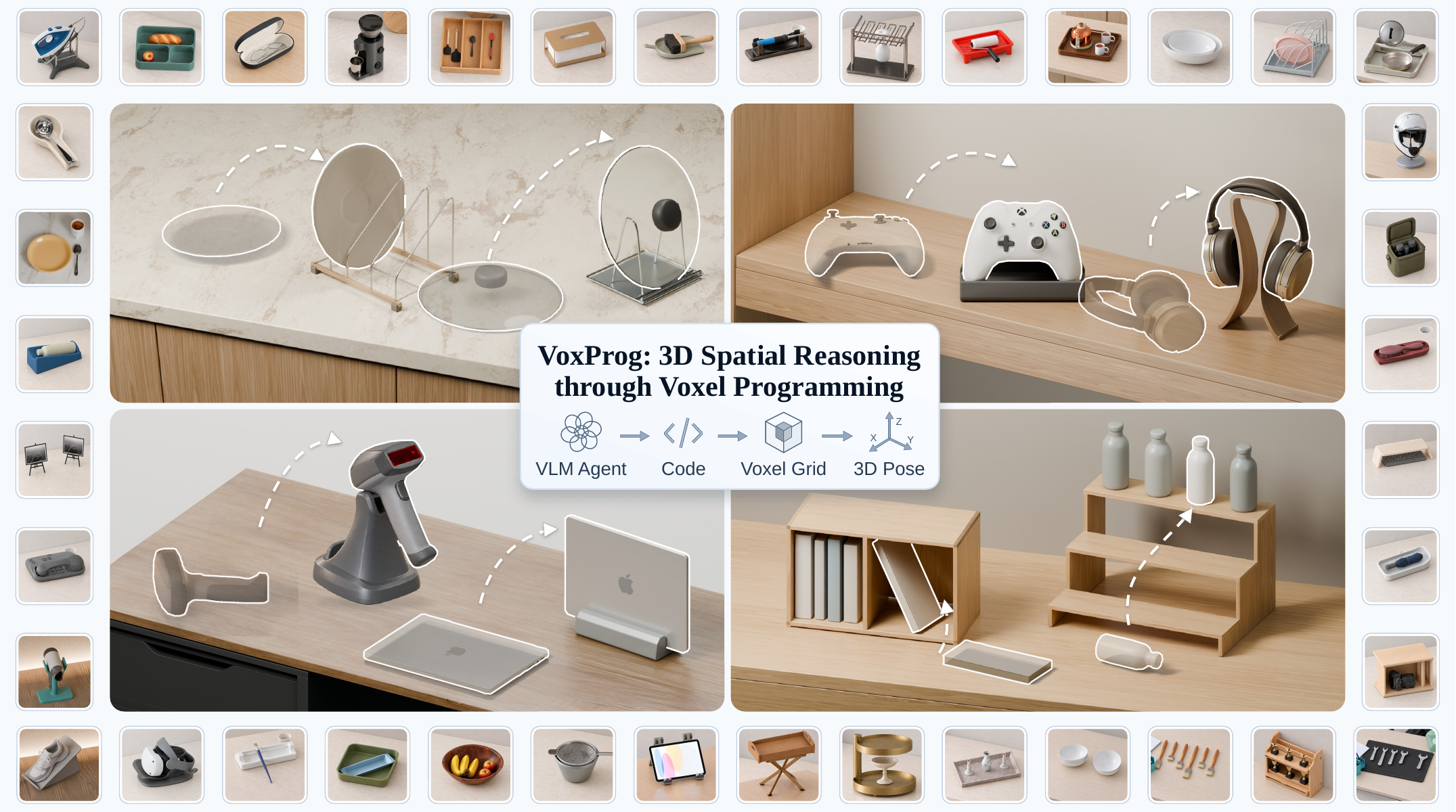}
\caption{\textbf{3D spatial reasoning through voxel programming.} Given a free-form spatial instruction, \method synthesizes executable Python programs that construct task-relevant surface references over local voxel views and compute target 6-DoF object placements. The surrounding scenes illustrate the diverse object geometries, part-level support structures, and multi-object spatial arrangements encountered in everyday placement tasks.}
\label{fig:teaser}
\end{figure}

\Needspace{10\baselineskip}
\section{Introduction}
\label{sec:intro}

Natural-language spatial instructions often specify goals whose geometric requirements must be inferred from the scene~\citep{robospatial2025,spatialvlm2024,spatialrgpt2024}. Placing a bottle inside an open display holder requires identifying the holder's receiving structure and determining how the bottle should be positioned and oriented to fit within it. Similarly, extending a row of cups requires inferring the direction and spacing of the existing arrangement to derive the next placement target. Here, the geometric references that define the goal are themselves outcomes of spatial computation over the 3D scene.

Existing approaches are limited by both the task requirements their interfaces can express and the geometric computations their representations support. Predefined task interfaces couple task coverage to task-specific implementations, limiting generalization to new spatial requirements~\citep{genmanip,innermonologue2022,saycan2022}. Geometric representations present a different difficulty: sparse keypoints can leave extended contact regions and local surface shapes under-specified~\citep{rekep2024,gensim2_2024,kpam2019,kite2023,moka2024}, while raw meshes require programs to organize irregular surface elements into the regions and references a task needs. Open-ended spatial reasoning therefore requires both programmatic computation that adapts to free-form instructions and a structured 3D surface representation that exposes scene geometry directly to code.

To this end, we formulate \emph{voxel programming}, in which a language model expresses its understanding of a spatial task as a program that constructs and operates on local voxel views (Fig.~\ref{fig:teaser}). By discretizing object surfaces onto regular grids at task-appropriate resolutions, local voxel views expose spatial coordinates and neighborhood relations directly as array operations. Generated programs can index these grids to isolate contact regions, extract boundaries, and measure local extent and orientation, turning a uniform grid representation into task-specific geometric references for 6-DoF pose computation.

\method realizes this formulation through a two-stage programming pipeline grounded by visual anchors and execution feedback. Starting from the instruction and raycast visual anchors, the model first writes an \emph{evidence program} to construct the surface regions required by the task, such as a dish rack's dividers and bottom support rails, and to specify the spatial relations they must satisfy. Conditioned on this constructed evidence, a \emph{pose program} then queries the selected geometry to compute the moved object's target pose. At both stages, numerical diagnostics and rendered views from program execution allow the model to verify intermediate geometry, revise faulty selections, and iteratively refine its generated code.

We evaluate \method on \suite, a new benchmark of 71 open-ended placement tasks where target poses depend on part-level affordances, 3D shape compatibility, and surrounding layouts. Because valid placements must satisfy both semantic intent and physical plausibility, our evaluation suite pairs a perceptual ranking metric, \emph{IA Rank Score}, with 3D \emph{Physical Validity} checks for static support and object interpenetration. IA Rank Score measures how consistently a method's rendered result outranks competitors in similarity to the ground-truth arrangement. Across \suite, \method substantially outperforms relation-based, keypoint-based, and mesh-based baselines, achieving its largest margin on geometric tasks where object contours govern fit and clearance. Ablation studies further show that constructing explicit geometric evidence before pose synthesis improves both physical validity and intent alignment over direct mesh programming, while local voxel views further improve intent alignment across all 71 tasks.

\noindent\textbf{Contributions.}
\ding{182}~We formulate voxel programming and develop \method, a two-stage system that constructs task-relevant surface geometry via local voxel views to compute 3D object placements.
\ding{183}~We introduce \suite, a benchmark of 71 language-guided 3D placement tasks spanning functional, geometric, and contextual arrangements.
\ding{184}~We propose a complementary evaluation protocol combining perceptual intent ranking (IA Rank Score) with 3D geometric verification (Physical Validity).

\section{Related Work}
\label{sec:related_work}

\par\Needspace{4\baselineskip}
\noindent\textbf{Spatial Task Specification.}
Language-guided placement has been studied through layout prediction, rearrangement search, and geometric optimization. Scene graphs encode inter-object relations for scene synthesis~\citep{graph2scene2025,commonscenes2023,instructscene2024,scenecraft2024}, while scene-layout and object-placement methods map language to spatial configurations~\citep{layoutgpt2023,holodeck2024,sceneteller2024,layoutvlm2025,fireplace2025,placeit3d2025,scenewiz3d2024,vulcan2025}. Rearrangement methods coordinate multi-object structures using learned spatial priors~\citep{structformer2022,structdiffusion2023,tang2024diffuscene,legonet2023,setitup2024} or pattern-guided search~\citep{lgmcts2024}. At a finer scale, object manipulation methods formulate 6-DoF pose optimization over sparse semantic keypoints, directional axes, and part-level affordances~\citep{kpam2019,gensim2_2024,rekep2024,sofar2025,omnimanip2025,pasg2025}.

While layout, surface-constraint, and keypoint formulations parameterize placement over bounding boxes, planar support patches, or sparse point and axis primitives, open-ended tasks often hinge on extended surface structures (such as receiving slots, angled ledges, curved rims, or recessed compartments) whose relevant regions depend on the instruction and paired objects. \method therefore treats geometric reference construction as an explicit, programmable stage of task specification: generated code composes spatial selections and neighborhood operations over local voxel views, retaining both the selected voxel regions and their clipped source-mesh patches for downstream pose measurement so that the same 3D object geometry can be segmented and parameterized differently across placement tasks.

\par\Needspace{4\baselineskip}
\noindent\textbf{Language Models for Spatial Computation.}
Executable programs let language models delegate quantitative calculation and visual queries to external interpreters~\citep{pal2023,pot2023,visprog2023,vipergpt2023,visualsketchpad2024}. In robotics, generated code sequences task plans~\citep{progprompt2023}, composes perception calls with control primitives~\citep{codeaspolicies2023,robocodex2024,chainofcode2024}, and synthesizes 3D value maps over workspace grids for motion planning~\citep{voxposer2023}. Recent methods further incorporate geometric processing via part-mask refinement~\citep{geomanip2025}, primitive fitting~\citep{embodiedcoder2025}, and frame-grounded tool calls~\citep{gca2026}.

In most programmatic frameworks, however, 3D perception remains encapsulated behind fixed API calls that return object centroids, bounding boxes, or pre-segmented masks~\citep{codeaspolicies2023,robocodex2024}, or operates on coarse workspace grids to score trajectories~\citep{voxposer2023}. We develop \emph{voxel programming} so that generated code can inspect and dissect 3D object surfaces directly at the part level. By discretizing candidate objects into local voxel views at task-chosen resolutions, \method turns 3D surface inspection into regular-grid array programming while retaining exact correspondence with the underlying mesh triangles. Generated code can thus construct custom surface witnesses on demand, verify them through visual and numerical execution feedback, and measure both discrete grid bounds and clipped source-mesh patches to solve for precise, physically valid 6-DoF object placements.

\FloatBarrier

\suppressfloats[t]
\begin{figure}[t]
\centering
\includegraphics[width=\linewidth]{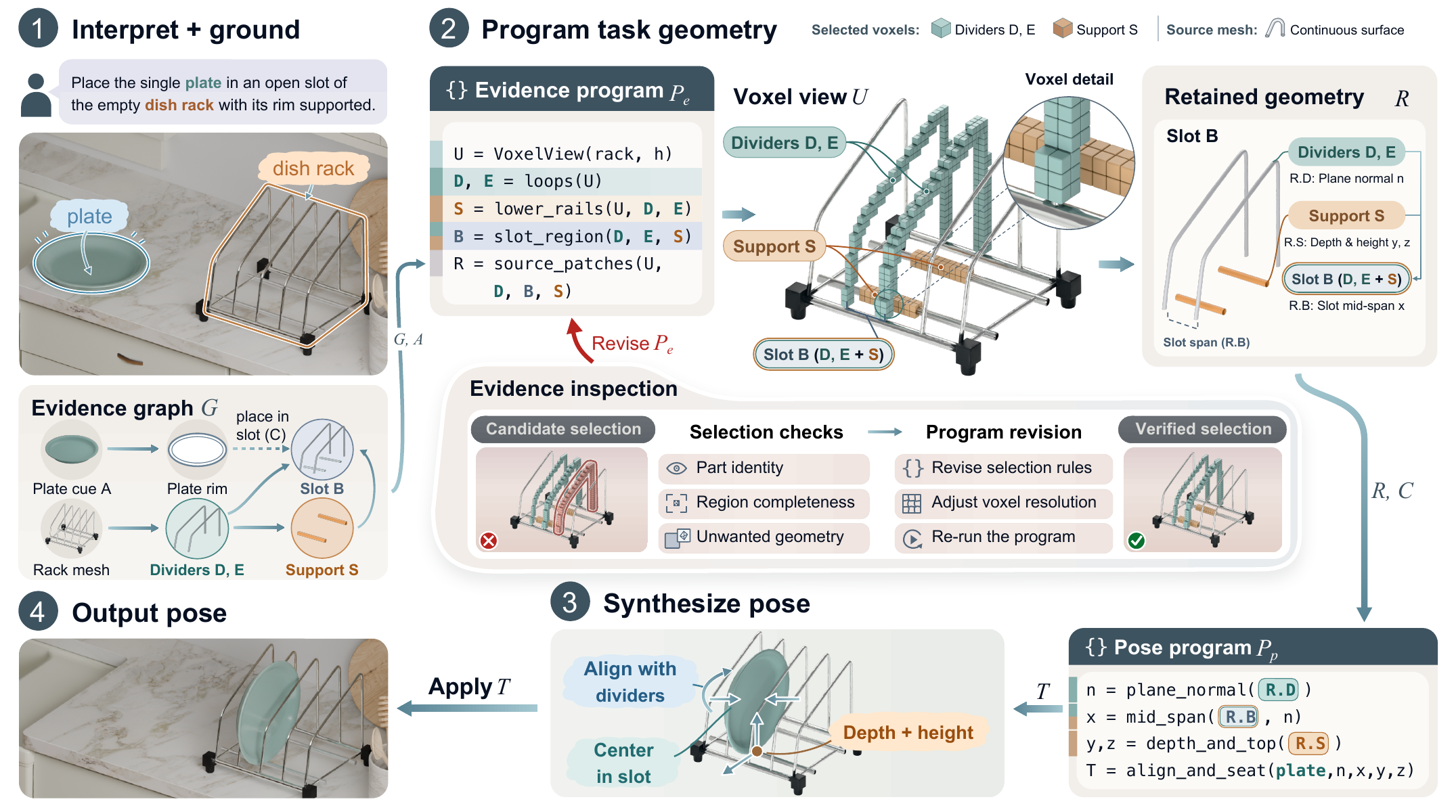}
\caption{\textbf{Overview of \method on upright plate placement.} (1)~The model parses the instruction into a directed evidence graph $\mathcal{G}$ and grounds visible cues into 3D surface anchors $\mathcal{A}$. (2)~An evidence program $P_{\mathrm{e}}$ operates on local voxel views to isolate the rack's dividers $D,E$, support rails $S$, and slot region $B$ in $\mathcal{R}$ along with relational constraints $\mathcal{C}$, refining its selections through visual and geometric checks. (3--4)~A pose program $P_{\mathrm{p}}$ queries the discrete voxel bounds and clipped source-mesh patches in $\mathcal{R}$ to solve for and apply the target pose $T\in SE(3)$.}
\label{fig:pipeline}
\end{figure}

\begingroup
\setlength{\parskip}{2pt}
\section{Method}
\label{sec:method}

Given a natural-language instruction $\mathcal{I}$ (e.g., \emph{``Place the single plate in an open slot of the empty dish rack with its rim supported''}) and a 3D scene $\mathcal{S}=\{o_n\}_{n=1}^{N}$ with known object meshes and initial poses, \method synthesizes and executes Python programs to compute a target placement pose $T\in SE(3)$. Rather than mapping language directly to 6-DoF coordinates in a single step, \method decouples \emph{geometric reference construction} from \emph{pose calculation}: an \emph{evidence program} $P_{\mathrm{e}}$ isolates task-relevant surface regions on the moved and reference objects, and a \emph{pose program} $P_{\mathrm{p}}$ measures these constructed surface regions and their relational constraints to solve for the moved object's final rigid transformation.

Fig.~\ref{fig:pipeline} illustrates this pipeline on an upright plate-placement task. The model first organizes the required spatial references into a directed evidence graph and grounds visible cues into 3D surface anchors (\S\ref{sec:contract}). Operating on local voxel views of scene meshes (\S\ref{sec:voxel}), the evidence program constructs the rack's adjacent dividers, lower support rails, and slot region while refining its selections through visual and geometric selection checks and program revision (\S\ref{sec:programming}). The verified evidence record and relational claims are then passed to the pose program, which queries both discrete voxel bounds and clipped source-mesh patches to align, center, and seat the plate upright within the target receiving slot (\S\ref{sec:interfaces}).

\subsection{Task Interpretation and Visual Grounding}
\label{sec:contract}

An LLM formalizes its understanding of the instruction as a directed evidence graph $\mathcal{G}=(\mathcal{V},\mathcal{E})$. Each node $v\in\mathcal{V}$ specifies either a visible semantic anchor or a derived geometric region to construct, along with its functional role in the task; each directed edge $(u,v)\in\mathcal{E}$ indicates that constructing $v$ depends on the location or geometry of $u$. In Fig.~\ref{fig:pipeline}, visual cues on the plate and dish rack initiate a dependency chain that extracts the plate's outer rim alongside the rack's adjacent dividers, support rails, and slot region.

Each node in $\mathcal{G}$ is routed according to how its geometry can be acquired. For a directly visible semantic cue, \texttt{Locate} queries a VLM for a 2D pixel coordinate in the rendered scene image and casts a camera ray against the scene mesh to recover a 3D surface point as a spatial anchor in $\mathcal{A}$. References that cannot be reduced to a single clickable pixel, such as an interior slot region, a recessed support ledge, or an occluded underside, are passed to the evidence program together with their descriptions and upstream 3D anchors, which localize where on the object mesh the program should inspect and slice the surface.

\subsection{Voxel Programming}
\label{sec:voxel}

Generated programs operate on scene geometry through local voxel views. Given a source mesh $s$ and voxel size $h>0$, $U=\texttt{VoxelView}(s,h)$ discretizes the surface of $s$ onto a regular 3D grid at resolution $h$. On this grid, the view exposes occupied surface cells as integer coordinate arrays $(N,3)$ alongside directional exposure masks and connected-component queries, while a rigid grid-to-world transform maps selected voxel indices and local bounds back to continuous 3D coordinates for downstream geometric computation.

Each constructed geometric reference forms a \emph{witness} in the evidence record $\mathcal{R}$. To construct a witness $v$, the evidence program instantiates a local view $U_v$ on source geometry $s_v$ at resolution $h_v$ and evaluates a programmatic binary mask $m_v$ over its surface cells to obtain the selected surface cells $W_v$:
\begin{equation}
U_v=\texttt{VoxelView}(s_v,h_v),\qquad
W_v=\{c\in U_v:m_v(c)=1\}.
\label{eq:voxel_witness}
\end{equation}
Because $U_v$ is a regular grid, the program can compose $m_v$ through standard array operations (axis-aligned slicing, proximity filters around visual anchors or upstream witnesses, directional exposure queries, and connected-component labeling) to isolate witnesses such as the rack's adjacent dividers $D,E$, lower support rails $S$, and combined slot region $B$ in Fig.~\ref{fig:pipeline}. For each witness $v$, the entry $\mathcal{R}.v$ in $\mathcal{R}$ stores these selected cells $W_v$, their parent view $U_v$, the source object $s_v$, and the provenance edges from $\mathcal{G}$.

Parameterizing the voxel size $h_v$ per view allows the evidence program to match discretization scale to local geometric structure without voxelizing the entire 3D scene at a uniform fine resolution. A program can use coarse grids to localize broad object surfaces and switch to a finer resolution $h_v$ when resolving thin wire dividers, narrow receiving slots, or shallow rims that would merge on a coarse grid, while retaining exact geometric correspondence with the underlying triangles of the source object mesh.

\Needspace{5\baselineskip}
\subsection{Evidence Program Generation and Refinement}
\label{sec:programming}

Conditioned on the instruction, evidence graph $\mathcal{G}$, scene $\mathcal{S}$, and grounded anchors $\mathcal{A}$, the LLM generates an executable evidence program $P_{\mathrm{e}}$ following the topological order of $\mathcal{G}$:
\begin{equation}
(\mathcal{R},\mathcal{C})=P_{\mathrm{e}}(\mathcal{S},\mathcal{A}),
\label{eq:evidence_program}
\end{equation}
where $\mathcal{R}$ stores the constructed witnesses and $\mathcal{C}$ specifies the target spatial relations among them (e.g., placing the plate's rim in slot $B$ between dividers $D,E$ while resting on support rails $S$ in Fig.~\ref{fig:pipeline}). Following the dependencies in $\mathcal{G}$ allows downstream witnesses to be defined relative to earlier selections.

To tie each witness to its functional role before writing code, the model first states explicit geometric inclusion and exclusion criteria for each target witness (such as excluding adjacent rack loops or vertical side walls). When a structural question remains unresolved, the model can execute lightweight \emph{probes} via the same voxel API to print numerical statistics (such as voxel counts, component counts, and coordinate spans) and inspect rendered diagnostic views of intermediate voxel selections.

Once a complete candidate $P_{\mathrm{e}}$ is executed, geometric checks and a VLM critic inspect each constructed witness for part identity, region completeness, and the absence of unwanted geometry (Fig.~\ref{fig:pipeline}, center). When a candidate selection leaks into adjacent structure or remains incomplete, this feedback guides the model to revise its programmatic selection rules or adjust the voxel resolution $h_v$ and re-run $P_{\mathrm{e}}$.

\subsection{Pose Program Generation and Refinement}
\label{sec:interfaces}

Given the verified evidence record $\mathcal{R}$ and relational specification $\mathcal{C}$, the LLM generates an executable pose program $P_{\mathrm{p}}$ to compute the moved object's target rigid transformation:
\begin{equation}
T=P_{\mathrm{p}}(\mathcal{R},\mathcal{S},T_{\mathrm{cur}}),\qquad T\in SE(3),
\label{eq:pose_program}
\end{equation}
where $T_{\mathrm{cur}}$ is the moved object's current world-frame pose (initialized at $T_0$) and $\mathcal{S}$ provides the scene meshes and initial poses. Every witness enters $P_{\mathrm{p}}$ as concrete geometry rather than pre-fitted point or vector constants, so $P_{\mathrm{p}}$ derives all required positions, axes, and offsets directly from $\mathcal{R}$ under $\mathcal{C}$.

Beyond discrete grid coordinates, each witness entry $\mathcal{R}.v$ in $\mathcal{R}$ also exposes the continuous source-mesh patch clipped to its selected voxels $W_v$, preserving exact face correspondence and outward surface normals. As shown in Fig.~\ref{fig:pipeline} (step 3), $P_{\mathrm{p}}$ combines discrete grid bounds with continuous mesh surfaces to solve accurately for the target pose $T$ on the original object geometry, computing the divider plane normal $\mathbf{n}$ from $\mathcal{R}.D$, the slot mid-span $x$ from $\mathcal{R}.B$, and the support depth and height $y,z$ from $\mathcal{R}.S$.

\Needspace{8\baselineskip}
Because all witnesses in $\mathcal{R}$ remain fixed in world coordinates at the initial scene state, $P_{\mathrm{p}}$ maps any initial world-space point $\mathbf{x}$ on the moved object (at initial pose $T_0$) to its position under a candidate pose $T$ as
\begin{equation}
\widetilde{\mathbf{x}}'=T T_0^{-1}\widetilde{\mathbf{x}},
\label{eq:pose_transform}
\end{equation}
where $\widetilde{\mathbf{x}}$ denotes homogeneous coordinates. After each execution of $P_{\mathrm{p}}$, the candidate pose $T$ updates $T_{\mathrm{cur}}$ in the 3D scene while $\mathcal{R}$ stays recorded in the initial scene state without recomputing evidence geometry, and the model receives a rendered observation of the resulting candidate placement together with numerical contact and clearance diagnostics to guide iterative revision of the pose program $P_{\mathrm{p}}$.

\par
\endgroup
\FloatBarrier

\suppressfloats[t]
\begingroup
\setlength{\parskip}{2pt}
\Needspace{6\baselineskip}
\section{Experiments}
\label{sec:experiments}

We evaluate \method on \suite against relation-, keypoint-, and mesh-based baselines (\S\ref{sec:setup}--\S\ref{sec:results}), ablate staged reference construction and local voxel views alongside external tasks from RoboCasa and Open6DOR v2 (\S\ref{sec:analyses}), and analyze physical and perceptual failure modes (\S\ref{sec:failure_analysis}).

\Needspace{5\baselineskip}
\subsection{Experimental Setup}
\label{sec:setup}

\begin{table}[t]
\centering
\begingroup\footnotesize
\definecolor{evalIA}{HTML}{649BC3}
\setlength{\tabcolsep}{3.8pt}\renewcommand{\arraystretch}{1.18}
\begin{tabular*}{\linewidth}{@{\extracolsep{\fill}}lcccc@{\hspace{13pt}}cccc@{}}
\toprule
& \multicolumn{4}{c}{\textbf{IA Rank Score} (\%) $\uparrow$} & \multicolumn{4}{c}{\textbf{Physical Validity} (\%) $\uparrow$}\\
\cmidrule(lr){2-5}\cmidrule(lr){6-9}
Method & Overall & Func. & Geom. & Context. & Overall & Func. & Geom. & Context.\\\midrule
\textsc{GenManip} & 25.59 & 25.00 & 32.14 & 22.99 & 73.2 & 67.9 & 92.9 & 69.0 \\
\textsc{GenSim2-kPAM} & 42.72 & 34.52 & 35.71 & \underline{54.02} & 77.5 & \underline{78.6} & 85.7 & 72.4 \\
\textsc{Direct Mesh} & \underline{48.59} & \underline{53.57} & \underline{44.05} & 45.98 & \underline{87.3} & 75.0 & \cellcolor{evalIA!12}\textbf{100.0} & \underline{93.1} \\
\midrule
\textbf{\method} & \cellcolor{evalIA!12}\textbf{83.10} & \cellcolor{evalIA!12}\textbf{86.90} & \cellcolor{evalIA!12}\textbf{88.10} & \cellcolor{evalIA!12}\textbf{77.01} & \cellcolor{evalIA!12}\textbf{95.8} & \cellcolor{evalIA!12}\textbf{92.9} & \cellcolor{evalIA!12}\textbf{100.0} & \cellcolor{evalIA!12}\textbf{96.6} \\
\bottomrule\end{tabular*}\par\smallskip
\endgroup

\caption{\textbf{Quantitative results on \suite.} We report IA Rank Score ($\uparrow$) and Physical Validity (\%, $\uparrow$) across all 71 placement tasks and within the Functional (28), Geometric (14), and Contextual (29) task categories. Best results in each column are highlighted in \textbf{bold} (shaded) and second-best results are \underline{underlined}.}
\label{tab:main_results}
\end{table}

\sssection{Benchmark.}
We evaluate on \suite, a benchmark of 71 individually curated language-guided 3D placement tasks. Each case presents an initial tabletop scene and a free-form instruction specifying the desired placement of an object. We emphasize diverse instructions that call for fine-grained reasoning about functional object parts, geometric fit, and contextual arrangements. Although an instruction can involve multiple spatial cues, we assign each case to one category according to its primary spatial requirement:
\begin{itemize}[leftmargin=*,itemsep=1pt,topsep=3pt,parsep=0pt]
 \item \textbf{Functional (28 cases):} The placement must support a specific use, such as placing a cup beneath a grinder chute to catch grounds, returning headphones to a stand, or orienting a funnel for pouring.
 \item \textbf{Geometric (14 cases):} The placement depends directly on object contours, such as nesting an oval bowl inside a matching container or fitting a baking tray into a narrow box at an angle.
 \item \textbf{Contextual (29 cases):} The placement takes its spatial cues from the surrounding multi-object arrangement, such as completing a missing place setting or restoring a symmetric tabletop layout.
\end{itemize}

Scenes are assembled from generated 3D assets through randomized placement and manual refinement. Detailed instructions, object geometry, and scene context make the intended arrangement well defined, with one manually annotated ground-truth 6-DoF pose per task. All methods receive the scene meshes, initial poses, instruction, and target-object identity; ground-truth poses and renderings serve only for evaluation.

\sssection{Evaluation Suite.}
We design an evaluation suite around two complementary requirements of object placement: realizing the intended spatial arrangement and maintaining physical plausibility. We evaluate Intent Alignment via visual comparison with ground truth and Physical Validity via 3D geometric checks.

\textit{Intent Alignment} measures perceptual similarity using DreamSim~\citep{dreamsim} between full-scene renderings of the predicted and ground-truth placements under matched cameras and lighting. Within each case, we rank the $M$ compared methods by increasing DreamSim distance and assign rank $r$ the score $100(M-r)/(M-1)$, using average ranks for ties. Averaging over cases gives the \textit{IA Rank Score} (0--100; higher is better), which measures how consistently a method outranks its competitors across tasks. Each task contributes equally; scores are compared within the same method pool.

\textit{Physical Validity} assesses whether the final placement establishes plausible static support and avoids interpenetration with surrounding scene objects. We report the percentage of evaluated task placements whose final 3D meshes satisfy both static support-contact and collision-clearance checks.

\sssection{Baselines.}
We compare \method against three language-guided placement approaches. \mbox{\textsc{GenManip}} \citep{genmanip} constructs placements using predefined spatial relations between target and anchor objects. \textsc{GenSim2-kPAM}~\citep{gensim2_2024,kpam2019} specifies placement targets through predicted semantic keypoints together with geometric point and axis constraints. \textsc{Direct Mesh} generates Blender Python code for placement and executes it through Blender-MCP\footnote{\url{https://github.com/ahujasid/blender-mcp}}.

\sssection{Model Configuration.}
All compared methods use GPT-5.6-Sol under a shared inference configuration for both language-guided code generation and visual grounding across all evaluated tasks.

\begin{figure}[!t]
\centering
\includegraphics[width=\linewidth]{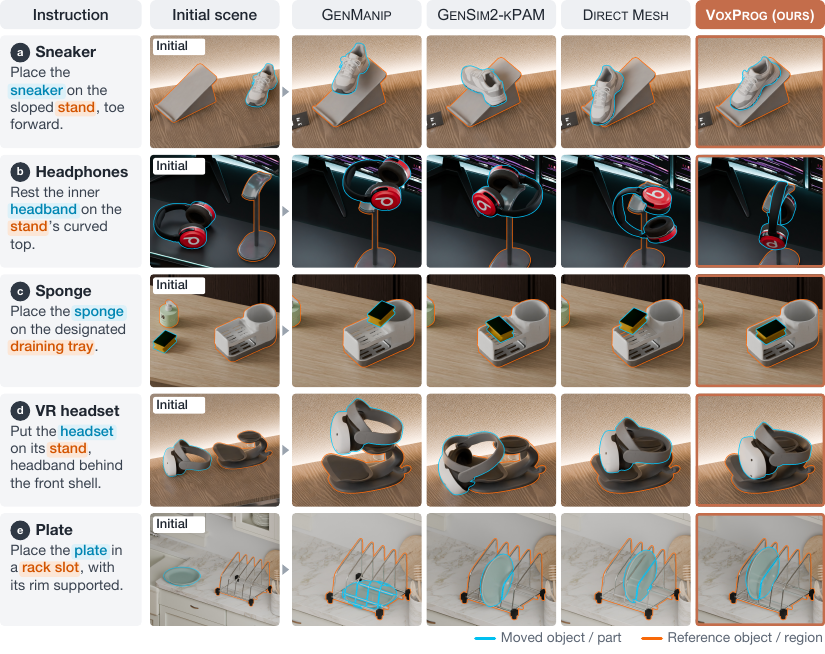}
\caption{\textbf{Qualitative comparisons on \suite.} Representative placements of a sneaker (a), headphones (b), a dish sponge (c), a VR headset (d), and a plate (e) across \textsc{GenManip}, \textsc{GenSim2-kPAM}, \textsc{Direct Mesh}, and \method. Colored highlights mark the \textcolor{movedobject}{moved object or part} and the \textcolor{referenceobject}{reference object or region} in both the instruction text and rendered scenes; terracotta borders highlight \method's final object placements.}
\label{fig:qualitative_diagnostics}
\end{figure}

\subsection{Results on the Task Suite}
\label{sec:results}
\sssection{Quantitative Results.}
\method achieves the highest IA Rank Score overall (\textbf{83.10}) and across all three task groups, alongside the highest Physical Validity at \textbf{95.8\%} (Table~\ref{tab:main_results}), outperforming \textsc{Direct Mesh} (48.59 / 87.3\%), \textsc{GenSim2-kPAM} (42.72 / 77.5\%), and \textsc{GenManip} (25.59 / 73.2\%). Although \textsc{Direct Mesh} also writes task-specific Python code over explicit scene meshes, \method improves its IA Rank Score by \textbf{34.51} points and Physical Validity by \textbf{8.5} points: code generation alone remains brittle without an intermediate representation that isolates the surfaces relevant to the instruction.

Breaking down performance by category reveals how different representations behave under distinct spatial demands. On Functional tasks, \method reaches \textbf{86.90} versus 53.57 for \textsc{Direct Mesh} and 34.52 for \textsc{GenSim2-kPAM} (with Physical Validity rising from 75.0\% and 78.6\% to \textbf{92.9\%}), as functional use requires isolating mating parts, such as a chute outlet or headphone arch, rather than whole-object bounding boxes or sparse keypoints. On Contextual tasks, the strongest baseline shifts from \textsc{Direct Mesh} (45.98) to \textsc{GenSim2-kPAM} (54.02), while \method retains a clear lead at \textbf{77.01}. The largest margin occurs on Geometric tasks: \method reaches an IA Rank Score of \textbf{88.10} versus 44.05 for \textsc{Direct Mesh}, although both attain \textbf{100.0\%} Physical Validity. The difference lies in resolving the local contours and clearances that distinguish the intended fit from other physically valid configurations.

\sssection{Qualitative Results.}
Fig.~\ref{fig:qualitative_diagnostics} exposes the gap between locating an object and identifying its functional contact region. In the headphone case (b), all three baselines bring the earcups against the stand or wrap the headphones around its post, missing the inner headband support specified by the instruction. \method matches the inner arch to the curved crown, establishing the intended part-level contact.

\Needspace{5\baselineskip}
On the sloped stand (a), \textsc{GenManip} leaves the sneaker above the support and \textsc{GenSim2-kPAM} turns it across the ramp. \method aligns the sole with the incline and keeps the toe facing forward. For the VR headset (d), \textsc{GenManip} and \textsc{GenSim2-kPAM} leave the headset above the cradle or in front of the stand, while \method positions it over the stand with the headband behind the front shell. Explicit surface references connect these local contact and orientation requirements to pose computation.

\Needspace{6\baselineskip}
\subsection{Ablation and External Evaluation}
\label{sec:analyses}

\sssection{Effect of Voxel Programming.}
To separate the effect of staged reference construction from that of local voxel views, we compare \textsc{Direct Mesh}, \textsc{Mesh Evidence}, and \method across all 71 tasks in Table~\ref{tab:ablation}. \textsc{Direct Mesh} writes a single placement program directly over the scene meshes; \textsc{Mesh Evidence} shares \method's two-stage pipeline and execution feedback but constructs witness regions through direct mesh operations on vertices and faces; \method constructs witnesses on local voxel views while retaining access to the clipped source-mesh surfaces during pose computation.

Within this three-method comparison, moving from single-stage \textsc{Direct Mesh} to the staged \textsc{Mesh Evidence} pipeline raises Physical Validity from 87.3\% to \textbf{95.8\%} and more than doubles the IA Rank Score from 26.76 to \textbf{56.34}. When a single program must simultaneously locate task-relevant surface geometry and compute 6-DoF coordinates over raw scene meshes, imprecise region selection directly degrades the resulting placement pose; introducing an explicit evidence stage allows candidate surface regions to be constructed, verified, and revised through execution feedback before downstream pose synthesis.

\par\smallskip
\noindent\begin{minipage}[t]{0.535\linewidth}
\vspace{0pt}
Replacing direct mesh selection with local voxel views raises the IA Rank Score from 56.34 to \textbf{66.90} while maintaining \textbf{95.8\%} Physical Validity. Regular grid indexing and neighborhood operations help resolve the intended receiving side of a structure. In Fig.~\ref{fig:ablation_qualitative}, \textsc{Mesh Evidence} places the tablet behind the stand's supports, which obstruct its screen, whereas \method seats it in front with the display unobstructed and facing the user.
\par

\sssection{External Placement Tasks.}
We evaluate generalization on 19 tasks adapted from RoboCasa~\citep{robocasa2024} and 13 from Open6DOR v2~\citep{open6dor2024,sofar2025}, formulated as final-state 3D placement problems under matched inputs (Fig.~\ref{fig:generalization}). On RoboCasa, where appliances and cabinets introduce semi-enclosed boundaries, \method achieves an IA Rank Score of \textbf{87.72} against 26.32--45.61 for the baselines, while maintaining \textbf{94.7\%} Physical Validity compared with 26.3\%--42.1\% for the baselines. On Open6DOR v2, \method leads in IA Rank Score (\textbf{61.54} vs.\ 57.69 for \textsc{Direct Mesh}) and raises Physical Validity to \textbf{92.3\%} versus 46.2\% for \textsc{Direct Mesh}: while broad directional instructions admit many 2D-plausible overhead positions, the baselines often leave objects floating or interpenetrating in 3D.
\end{minipage}\hfill
\begin{minipage}[t]{0.45\linewidth}
\vspace{0pt}
\captionsetup{type=figure}
\centering
{\captionsetup{type=table}
\begingroup
\definecolor{evalIA}{HTML}{649BC3}
\setlength{\tabcolsep}{3pt}
\renewcommand{\arraystretch}{1.16}
\footnotesize
\begin{tabular*}{\linewidth}{@{\extracolsep{\fill}}lcc@{}}
\toprule
Method & \begin{tabular}[c]{@{}c@{}}IA Rank\\Score (\%) $\uparrow$\end{tabular} & \begin{tabular}[c]{@{}c@{}}Physical\\Validity (\%) $\uparrow$\end{tabular}\\
\midrule
\textsc{Direct Mesh} & 26.76 & 87.3 \\
\textsc{Mesh Evidence} & \underline{56.34} & \cellcolor{evalIA!12}\textbf{95.8} \\
\textbf{\method} & \cellcolor{evalIA!12}\textbf{66.90} & \cellcolor{evalIA!12}\textbf{95.8} \\
\bottomrule
\end{tabular*}
\endgroup

\caption{\textbf{Ablation of voxel programming.} Results across all 71 placement tasks on \suite.}
\label{tab:ablation}}
\vspace{7pt}
\includegraphics[width=\linewidth]{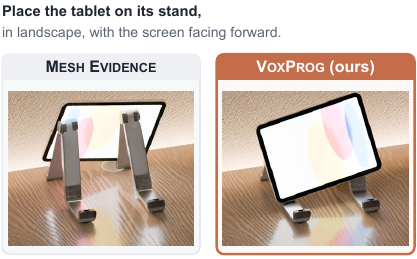}
\caption{\textbf{Tablet placement ablation.} The supports obstruct the screen in \textsc{Mesh Evidence}; \method keeps the display clear and facing forward.}
\label{fig:ablation_qualitative}
\end{minipage}
\par\smallskip
\begin{figure}[!htbp]
\centering
\includegraphics[width=\linewidth]{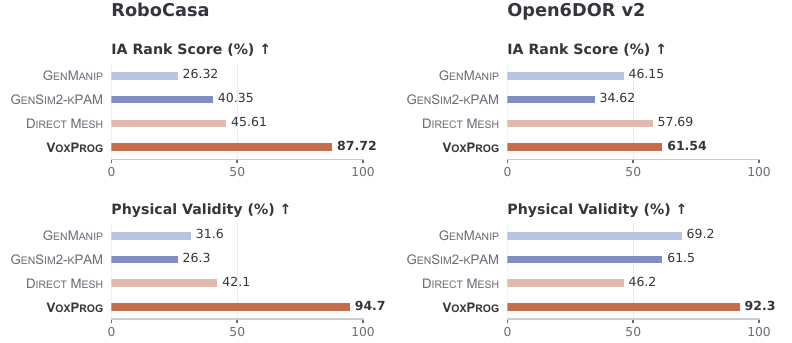}
\caption{\textbf{External benchmark evaluation.} IA Rank Score ($\uparrow$) and Physical Validity ($\uparrow$) on 19 kitchen tasks adapted from RoboCasa and 13 tabletop placement tasks from Open6DOR v2 across all four compared methods.}
\label{fig:generalization}
\end{figure}

In the mug task (Fig.~\ref{fig:external_qualitative}a), \textsc{GenManip} leaves the mug beside the dispenser, while \textsc{GenSim2-kPAM} and \textsc{Direct Mesh} leave it interpenetrating the drip tray. \method centers the opening beneath the nozzle and rests the base on the tray, jointly satisfying the alignment and support required to receive coffee. The SD-card task (c) exposes a complementary contact failure: \textsc{GenManip} and \textsc{Direct Mesh} leave the card floating above the USB drive, whereas \method seats it on the top surface.
\par\smallskip
\begin{figure}[t]
\centering
\includegraphics[width=\linewidth]{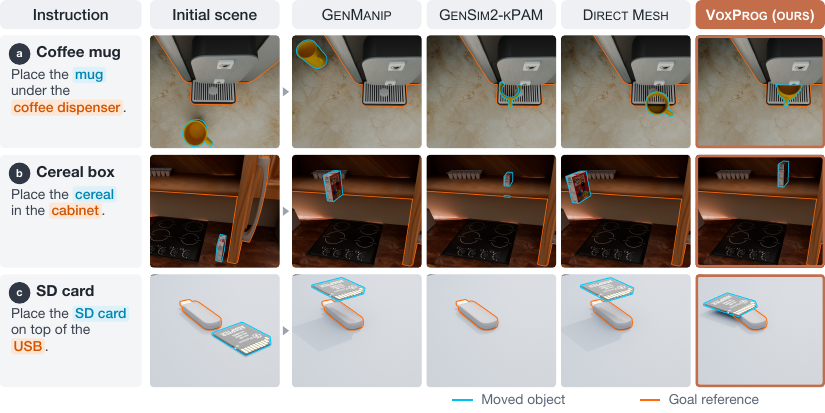}
\caption{\textbf{Qualitative results on external benchmarks.} Mug (a) and cereal-box (b) placement from RoboCasa, and SD-card placement (c) from Open6DOR v2, comparing \textsc{GenManip}, \textsc{GenSim2-kPAM}, \textsc{Direct Mesh}, and \method. Colored contours mark the \textcolor{movedobject}{moved object} and \textcolor{referenceobject}{goal reference}; terracotta borders highlight \method.}
\label{fig:external_qualitative}
\end{figure}

\Needspace{4\baselineskip}
\subsection{Failure Analysis}
\label{sec:failure_analysis}

\noindent\textbf{Physical Failure Profiles.} Fig.~\ref{fig:failure_diagnostics}a decomposes physical rejections on \suite into support and interpenetration failures. \textsc{GenManip} incurs 19 rejections (11 support, 8 intersection), versus 16 for \textsc{GenSim2-kPAM} (13 intersection, 3 support). These patterns are consistent with coarse relation offsets leaving support contact under-specified and sparse point-and-axis constraints leaving surface clearance under-constrained. \textsc{Direct Mesh} has 9 rejections (6 intersection, 3 support), while both staged variants cut rejections to 3 (3 intersection for \textsc{Mesh Evidence}; 2 intersection, 1 support for \method).

\setlength{\intextsep}{0pt}
\begin{wrapfigure}[25]{r}{0.415\linewidth}
\centering
\setlength{\abovecaptionskip}{2pt}
\includegraphics[width=\linewidth]{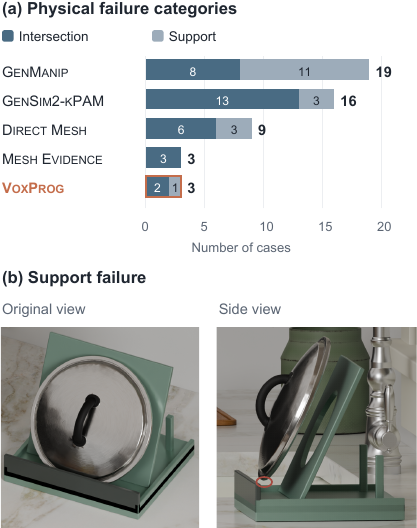}
\caption{\textbf{Failure profiles and contact checks.} (a) Physical rejections on \suite. (b) A lid seated from the front floats from the side.}
\label{fig:failure_diagnostics}
\label{fig:pot_lid_contact}
\end{wrapfigure}
\noindent\textbf{Visually Ambiguous Contact.} The pot-lid case in Fig.~\ref{fig:failure_diagnostics}b illustrates why 2D visual inspection alone cannot establish physical validity in 3D object placement. From the primary camera view, the lid appears properly seated: it stands upright with its handle facing outward, its circular contour centered between the side supports, and its lower rim aligned over the receiving rack. A side view of the same final pose reveals, however, that the lid remains suspended above the rack, separated by a clear vertical gap from both the base tray and the inclined backrest. Here, the rack's opaque front frame provides a strong visual cue to the destination while occluding the recessed region where support contact should occur. Because 2D projection overlap along the viewing ray can conceal 3D separation under self-occlusion, a single-view visual comparison can accept a physically unsupported placement even when the target object and its apparent relation to the reference structure look aligned in image space. Pairing perceptual evaluation with 3D geometric verification addresses this ambiguity by checking local 3D support contact and interpenetration conditions between the placed object and surrounding scene meshes, making the distinction between apparent 2D seating in the rendered camera view and actual physical support in the 3D scene explicit.
\par
\WFclear

\FloatBarrier

\enlargethispage{\baselineskip}
\vspace{-30pt}
\section{Conclusion}
\label{sec:conclusion}

We presented \method, a framework for 3D spatial reasoning that formulates language-guided object placement through \emph{voxel programming}. Rather than parameterizing tasks over fixed spatial relations or sparse keypoint primitives, \method treats the geometric references required by an instruction as intermediate surface structures constructed by executable code. By coupling regular-grid selection over local voxel views with exact surface normals and patches from the underlying meshes, our two-stage pipeline enables a language model to isolate functional contact regions, verify them through visual and geometric feedback, and solve for physically valid $SE(3)$ poses. Across the 71-task \suite benchmark and external tasks from RoboCasa and Open6DOR v2, \method consistently outperforms relation-based, keypoint-based, and mesh-based baselines in both intent alignment and physical validity. Coupling such programmable surface witnesses with downstream motion generation provides a natural direction for extending fine-grained geometric reasoning from target placement specification to full manipulation execution.\looseness=-1

\par
\endgroup

\clearpage
\bibliography{main}
\bibliographystyle{references}

\end{document}